\documentclass{bmvc2k}
\usepackage{amsmath}

\title{Low Light Video Enhancement by Learning on Static Videos with Cross-Frame Attention}

\addauthor{Shivam Chhirolya}{shivamchhirolya@gmail.com}{1}
\addauthor{Sameer Malik}{sameer@iisc.ac.in}{1}
\addauthor{Rajiv Soundararajan}{rajivs@iisc.ac.in}{1}

\addinstitution{
 Department of Electrical Communication Engineering,\\
 Indian Institute of Science,\\
 Bangalore, India\\
}

\runninghead{Shivam, Sameer, Rajiv}{LLVE BY LEARNING ON STATIC VIDEOS WITH CFA}

\begin{document}

\maketitle

\begin{abstract}
The design of deep learning methods for low light video enhancement remains a challenging problem owing to the difficulty in capturing low light and ground truth video pairs. This is particularly hard in the context of dynamic scenes or moving cameras where a long exposure ground truth cannot be captured. We approach this problem by training a model on static videos such that the model can generalize to dynamic videos. Existing methods adopting this approach operate frame by frame and do not exploit the relationships among neighbouring frames. 
We overcome this limitation through a self-cross dilated attention module that can effectively learn to use information from neighbouring frames even when dynamics between the frames are different during training and test times.
We validate our approach through experiments on multiple datasets and show that our method outperforms other state-of-the-art video enhancement algorithms when trained only on static videos.
\end{abstract}


\label{chap:Introduction}
\section{Introduction}
\label{sec:intro}
Camera captured videos under low light conditions often suffer from poor contrast and noise due to the limited exposure time allowed by the typical frame rates. While convolutional neural network (CNN) models have been very successful in low light image enhancement tasks \cite{DRBN, uretinexnet, DCCNet, DLN}, low light video enhancement still remains challenging due to a lack of real world datasets with pairs of low light and ground truth videos. This is because it is particularly hard to capture labelled video pairs for dynamic scenes and has been a serious limitation in the use of CNN models for low light video enhancement. We thus focus on the problem of training video enhancement on static videos such that they can be directly applied on dynamic videos. 

A naive approach such as training just on individual video frames may result in temporally inconsistent enhancement which may manifest as visually displeasing flickering artefact. Chen et al. \cite{SMID} address this challenge by training a Siamese model on static videos that enforces consistent output among all the video frames \cite{SMID}. They show that the Siamese model achieves more temporally consistent enhancement across frames with reduced flickering artefacts. However, their approach does not exploit the highly correlated information in neighbouring frames for effective video enhancement.

While there exist other video restoration methods that  effectively use information from neighbouring frames \cite{VRT,FastDVDnet}, they mostly rely on jointly training an optical flow estimation and a video restoration model. However, an optical flow model trained on a static video dataset may not generalize well to real world dynamic videos. An alternate approach to make use of neighboring frames during enhancement is to synthetically generate distorted videos \cite{MBLLEN, CycleGAN, Learning_temporal}. However, inaccuracies in distortion modelling and data generation may result in sub-optimal performance of the video restoration models trained on such datasets.  

To address these challenges, we develop a video enhancement model that uses information from neighbouring frames, achieves consistent restoration across frames and yet generalizes well to dynamic real videos even when trained on static videos. Specifically, we design a novel video enhancement model that uses cross-attention to exploit information from neighbouring frames to achieve high quality and temporally consistent video restoration. Since the cross-attention module explicitly computes similarity with features from neighbouring frames, it can generalize better to real dynamic videos even when trained only on static videos. 

The computation of cross-attention is computationally expensive, especially when it is computed for the whole of the neighbouring frame corresponding to every given pixel in the reference frame. Thus we compute cross-attention only in a spatial neighbourhood of a given reference frame pixel in the neighboring frame. However, this limits its usefulness in the case of large motions where the reference frame pixel may not be present in the local spatial neighbourhood. To mitigate this, we augment cross-attention with dilated cross-attention that enlarges the spatial neighbourhood while retaining the computational effort. We incorporate these novel components into a multi-scale architecture with blocked attention and self-attention. We note that one of our main contributions is the use of self-cross attention to enable the method to work on dynamic videos despite being trained only on static videos. Further, our use of self-cross attention differs from VRT \cite{VRT} in our goal to get rid of explicit optical flow altogether for video enhancement in contrast to its use to improve motion estimation in VRT.  

Overall the main contributions of this work are:
\begin{itemize}
\item	The novel use of a cross-attention module that exploits inter-frame interactions for superior enhancement of dynamic videos despite the model being trained only on static videos. 

\item The use of dilated cross-attention for effective enhancement in videos with large motion. 

\item The creation of a novel dynamic low light video dataset that consists of real world distortions with synthetic motion for performance evaluation. 

\item Superior objective or subjective performance on multiple datasets of dynamic low-light videos when compared to other methods also trained on static videos.   
\end{itemize}

Note that our main contribution is in training a model that uses neighboring frames on static videos and enables it to generalize well for dynamic videos. Although SMID \cite{SMID} is also trained on static videos and applied on dynamic videos, it only works with individual frames at test time and does not use neighboring frames. This is one the main reasons why our method achieves superior performance when compared to SMID. 
\label{chap:review}
\section{Related Work}
We survey related work in the areas of low light video enhancement, transformer based methods for video restoration and low light image enhancement. 

\textbf{Low Light Video Enhancement:} One of the earliest approaches for deep video enhancement involves replacing 2D convolutions with 3D convolutions in a method designed for image enhancement \cite{MBLLEN}. This approach relies on the availability of paired ground truth and low light dynamic videos. This method suffers from limitations when trained on static and tested on dynamic videos. Since then, researchers have tried a variety of approaches to account for the lack of ground truth data for dynamic videos. Chen et al. \cite{SMID} were successful in enhancing low light videos by training on static scenes through the imposition of a temporal consistency loss on different frame output. However, this method does not exploit the correlations in neighboring frames. Zhang et al. \cite{Learning_temporal} synthesize motion in single images through segmentation and generation of optical flow vectors to train on a synthetic dataset. SIDGAN uses a CycleGAN based approach to generate paired low light and ground truth videos \cite{CycleGAN}. However, it may be less cumbersome to design methods where such an intermediate data generation step is not necessary. Alternately, an optical system was designed to help capture pairs of low light and high quality ground truth videos \cite{Learning_to_see_moving_objects_in_the_dark}. 

\textbf{Video Restoration:} Tassano et al. \cite{FastDVDnet} proposed FastDVDnet, which contains a two-layered ResUnet architecture to exploit the correlation among neighboring frames without explicitly computing optical flow for video denoising. 
Attention mechanisms through transformer based architectures have also attracted a lot of attention in video restoration in conjunction with convolutional neural networks \cite{VideoSuperResolutionTransformer, VRT, VideoRestorationWithEnhancedDeformableConvolutionalNetworks}. Wang et al. \cite{VideoRestorationWithEnhancedDeformableConvolutionalNetworks} learn pixel-level attention maps for spatial and temporal feature fusion.
Cao et al. \cite{VideoSuperResolutionTransformer} propose to use self-attention among local patches within a video. Jingyun et al. \cite{VRT} propose a self-cross attention and optical flow based video restoration transformer (VRT). Nevertheless, these architectures have not been explored for their relevance in low light video enhancement. Researchers have also explored various methods to enforce temporal consistency in video restoration such as those based on achieving consistency with geometric transforms \cite{Single_frame_regularization_for_temporally_stable_cnns} as well as the use of long short term memory units \cite{Learning_blind_video_temporal_consistency}.

\textbf{Low Light Image Enhancement:} Most deep image enhancement architectures either use the retinex model \cite{uretinexnet, retinex_unroll, RetinexDIP} or multi-scale subband processing \cite{LPNet, DSLR, DRBN}. There also exist some end-to-end learning approaches such as MBLLEN \cite{MBLLEN} or DLN \cite{DLN}. Nevertheless, these methods are not effective for video enhancement as they do not have any mechanism to ensure temporal consistency in the enhanced videos.   




\label{chap:proposed}
\section{Overall Framework}
We first present a base model based on several successful elements of transformer based architectures for image processing. We then incorporate our contributions in cross-frame processing in this set up. We start with a base model consisting of a multi-scale architecture where the processing in each scale consists of self-attention \cite{Attnetion} and feature blocking \cite{HITGAN} components. To enable inter-frame processing, we modify the base model by introducing a cross-attention feature extraction module to allow for interaction between the neighbouring frame features. In the following, we first briefly discuss the base model and then present our cross-attention module and other modifications to the base model.

\subsection{Base Model}\label{Base-Model}
\begin{figure}
\begin{tabular}{cc}
\bmvaHangBox{{\hspace{+1cm}\includegraphics[width=5.5cm]{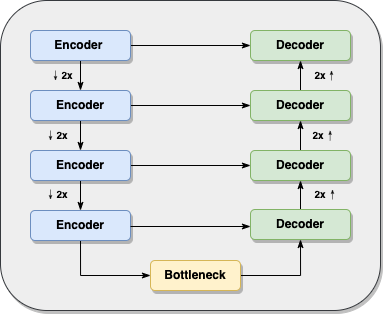}}}&

\bmvaHangBox{{\hspace{+1cm}\includegraphics[width=3.2cm]{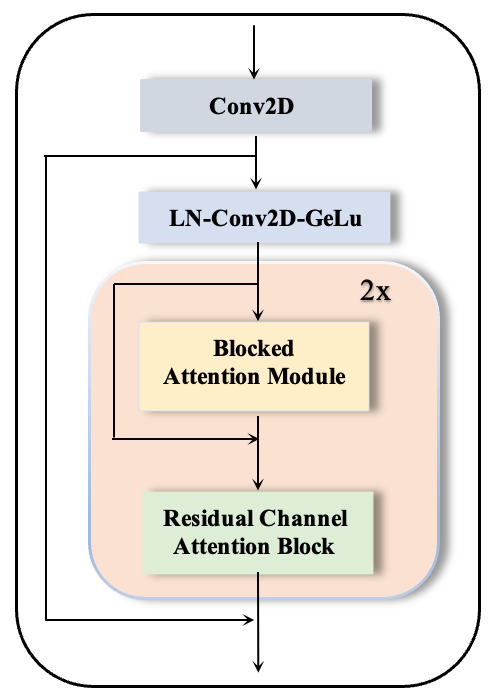}}}\\
\hspace{+1cm}(a)&\hspace{+1cm}(b)
\end{tabular}
\caption{(a) Overall framework of the proposed method , (b) Encoder/Decoder/Bottleneck architecture. For Decoder, attention module is Base Blocked Attention Module illustrated in Figure \ref{BaseAttnModule}. For Encoder \& Bottleneck, Blocked Attention module is illustrated in Figure \ref{EncAttnModule}. LN represents Layer-norm.}
\label{fig:Arch}
\end{figure}

The base model framework follows the popular encoder-decoder design in UNet \cite{Unet} as shown in Figure \ref{fig:Arch}a. The architecture consists of encoders, decoders and a bottleneck. We first describe the encoder processing now. 
Let the input to Encoder $i$, $i\in\{1,2,\dots,M\}$, be $\mathbf{f}_i^t$ of dimension $H_i\times W_i\times C_i$ where $H_i, W_i$ and $C_i$ denote the height, width and number of channels respectively and $t$ corresponds to the frame index from which the features are obtained. 
Each of the encoders includes initial processing using two convolutional layers, a normalization layer and a GeLU non-linearity as shown in Figure \ref{fig:Arch}b. We then block the resulting features into a tensor of shape $(b\times b, H_i/b \times W_i/b, C_i)$. This essentially partitions the features into non-overlapping blocks of size $b\times b$ as shown in Figure \ref{BaseAttnModule}. Each of the blocks referred to as $\mathbf{f}_{ij}^{t}$, where $j\in\{1,2,\ldots,\frac{H_i}{b}\times \frac{W_i}{b}\}$ are then processed using a multi-headed self-attention block (MHSA) \cite{MHA}. We note that  $\mathbf{f}_{ij}^{t}$ is flattened spatially to obtain a dimension $b^2\times C_i$ . We describe the MHSA block as follows.

MHSA with $L_i$ heads first linearly projects $\mathbf{f}_{ij}^{t}$ using $C_i\times C_i$ matrices $P^Q_{ih}$, $P^K_{ih}$ and $P^V_{ih}$ to compute query $Q_{ijh}$, key $K_{ijh}$ and value $V_{ijh}$ respectively, where $h\in\{1,2,\ldots,L_i\}$ and each is of dimension $b^2\times C_i$. 
Note that the linear projection matrices are shared across all the blocks for a given encoder. 
Now we compute an attention map for each head $h\in \{1,...,L_i\}$  using $Q_{ijh}^t$ to query the key $K_{ijh}^t$ as
\begin{equation}\label{eqn:self-attention}
SA_{ijh}(t)=\left[\operatorname{SoftMax}\left(Q_{ijh}^t\left(K_{ijh}^t)\right)^{T} / \sqrt{C_i}\right)\right]V_{ijh}^t,
\end{equation}
where softmax is performed rowwise. 
Finally we get MSHA of the block by concatenating $SA^{ijh}(t)$ across all the heads.

We then unblock by rearranging the features of all blocks to get features of dimension $(H_i, W_i, C_i)$. We refer to the unblocked features as $SA_{i}(t)$. We further process the features using residual channel attention block \cite{RCAB} and downsample them to get the features $\mathbf{f}_{i+1}^t$ from the Encoder $i$. We employ the same architecture as the encoder for the decoder and the bottleneck in the Base Model.

\begin{figure}[t]
\centering
\bmvaHangBox{{\includegraphics[width=12.4cm]{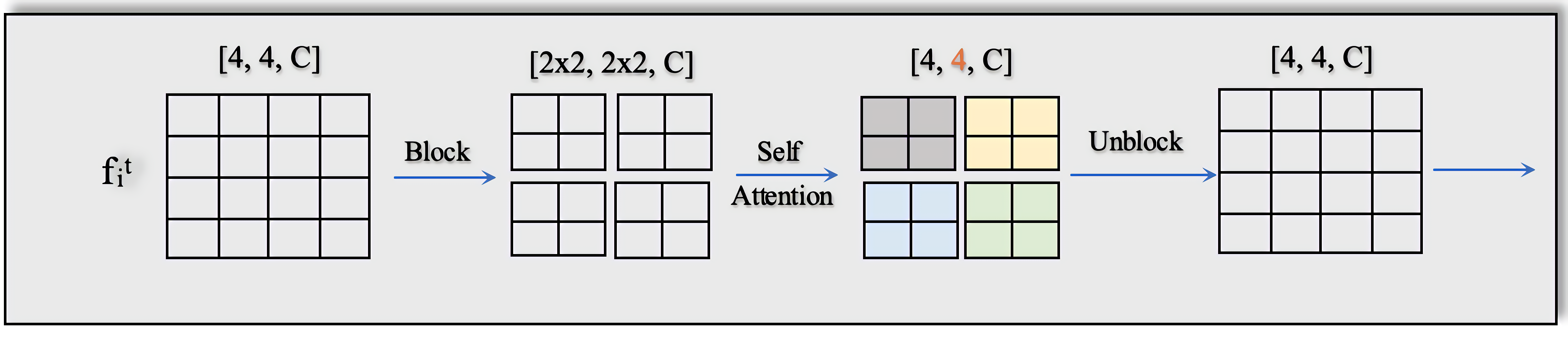}}}
\caption{Base Blocked Attention Module. This figure explains the process from blocking to unblocking}
\label{BaseAttnModule}
\end{figure}

\subsection{Proposed Dual Self-Cross Attention Module}
\label{sec:dual_attention}
To introduce inter-frame interactions into our base model, we use cross-attention between feature maps of two neighboring frames. This cross-attention is computed in addition to the self-attention for each of the frames as described in Equation (\ref{eqn:self-attention}). We refer to our module as a dual self-cross attention feature extraction module. Specifically, to process a given frame $\mathbf{y}_t$, our model also takes past and future frames $\{\mathbf{y}_{t-1},  \mathbf{y}_{t+1}\}$ in addition to the current frame $\mathbf{y}_t$ as input. Then, Encoder $i$ takes three sets of features $\{\mathbf{f}^{t-1}_i, \mathbf{f}^{t}_i, \mathbf{f}^{t+1}_i\}$ as input and processes them  using self and cross attention followed by feature fusion to produce $\{\mathbf{f}^{t-1}_{i+1}, \mathbf{f}^{t}_{i+1}, \mathbf{f}^{t+1}_{i+1}\}$. We first describe the cross-attention computation for any two frames and then describe the all the computations for three neighboring frames in our dual self-cross attention module. 

\begin{figure}
\begin{tabular}{cc}
\bmvaHangBox{{\includegraphics[width=10cm]{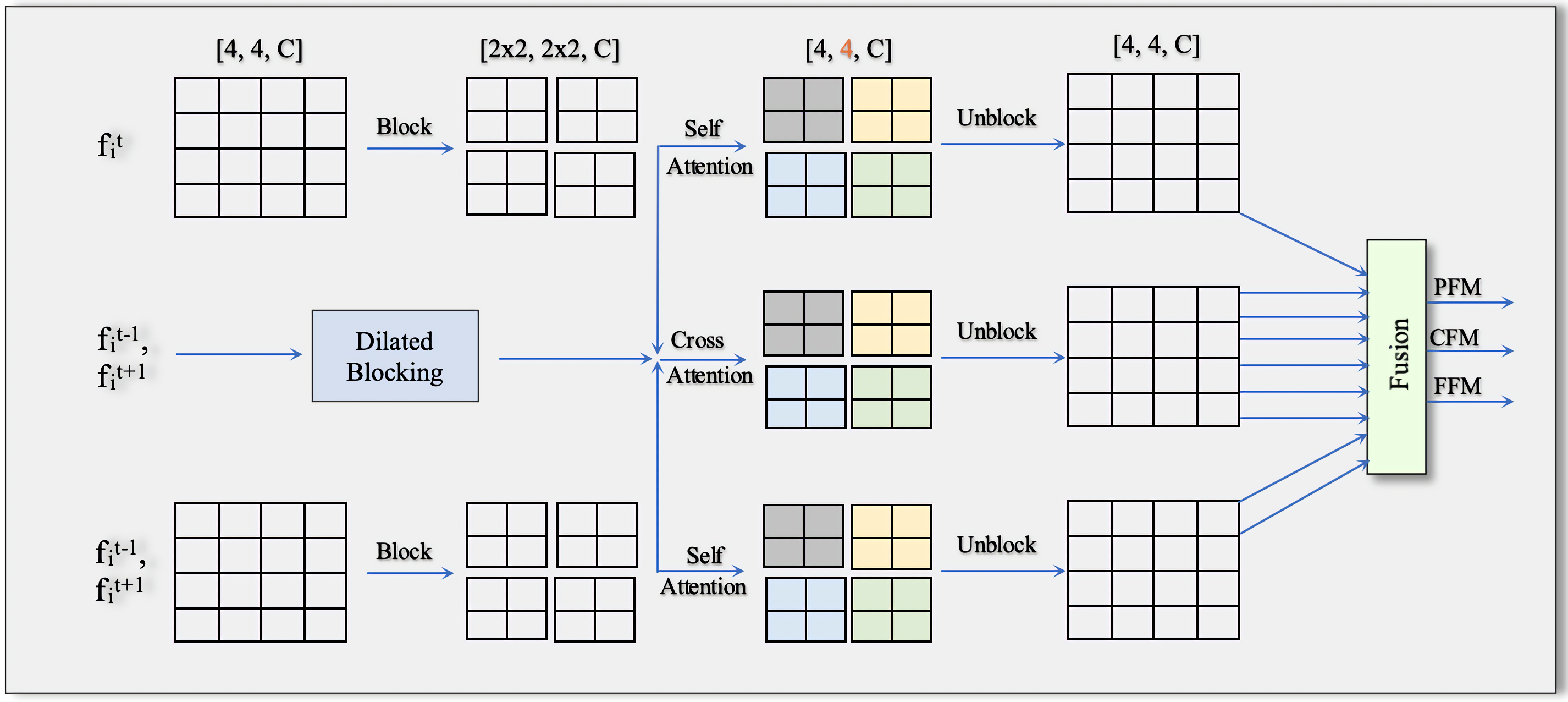}}}&
\bmvaHangBox{\hspace{-3mm}\fbox{\includegraphics[width=2.38cm]{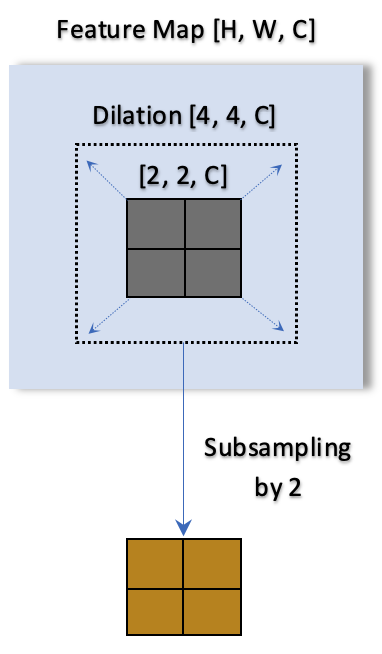}}}\\
(a)&(b)
\end{tabular}


\caption{(a) Encoder/Bottleneck Attention Module. We perform blocking for $\{f^{t-1}_i, f^{t}_i, f^{t+1}_i\}$ and dilated blocking for $\{f^{t-1}_i, f^{t+1}_i\}$, which we use to compute three self attention maps and six cross \& dilated cross attention maps.
(b) Dilated Blocking. To get a dilated block of size $b \times b$, we divide the given feature map into $2b \times 2b$  overlapping blocks and then subsample these blocks by a factor of 2.
For better understanding on these attention maps, see Figure \ref{fig:totalattn}. We fuse these maps, to get past feature map (PFM), current feature map (CFM) and future feature map (FFM).}
\label{EncAttnModule}
\end{figure}

\textbf{Cross Attention:} Consider features $\mathbf{f}^{t-1}_i$ and $\mathbf{f}^{t}_i$. To process $\mathbf{f}^{t}_i$ using cross-attention with $\mathbf{f}^{t-1}_i$, we first block them as described in Figure \ref{EncAttnModule} to obtain $\mathbf{f}_{ij}^{t-1}$ and $\mathbf{f}_{ij}^{t}$, where $j\in\{1,2,\ldots,\frac{H_i}{b}\times \frac{W_i}{b}\}$. We then linearly project $\mathbf{f}_{ij}^{t-1}$ using the matrices $P_{ih}^K$ and $P_{ih}^V$ to compute key $K_{ijh}^{t-1}$ and value $V_{ijh}^{t-1}$ respectively for head $h$. We also project features $\mathbf{f}^{t}_{ij}$ using $P_{ih}^Q$ to compute query $Q_{ijh}^{t}$. We now compute the cross-attention output between features in frame $t$ and $t-1$ as 

\begin{equation}\label{eqn:cross-attention}
CA_{ijh}(t,t-1)=\left[\operatorname{SoftMax}\left(Q_{ijh}^t\left(K_{ijh}^{t-1}\right)^{T} / \sqrt{C_i}\right)\right]V_{ijh}^{t-1},
\end{equation}
where we compute SoftMax in row-wise fashion as before. We concatenate the features from all the heads to obtain $CA_{ij}(t,t-1)$. Note that by reversing the roles of $t$ and $t-1$, we obtain $CA_{ij}(t-1,t)$ which is a processed version of $\mathbf{f}_i^{t-1}$. 

$CA_{ij}(t,t-1)$ computes a cross-attention between blocks centered at the same location across frames. 
However, in case of large motion, many of the pixels in $\mathbf{f}_{ij}^{t}$ may not present in $\mathbf{f}_{ij}^{t-1}$. To address this, we further compute cross-attention between $\mathbf{f}_{ij}^{t}$ and a dilated version of $\mathbf{f}_{ij}^{t-1}$. We denote this as $DCA_{ij}(t,t-1)$. Figure \ref{EncAttnModule} explains the method to get the dilated version of a given feature map. The feature maps for all the blocks are then unblocked to obtain $CA_i(t,t-1)$ and $DCA_{i}(t,t-1)$.

\begin{figure}[t]
\centering
\bmvaHangBox{{\includegraphics[width=12cm]{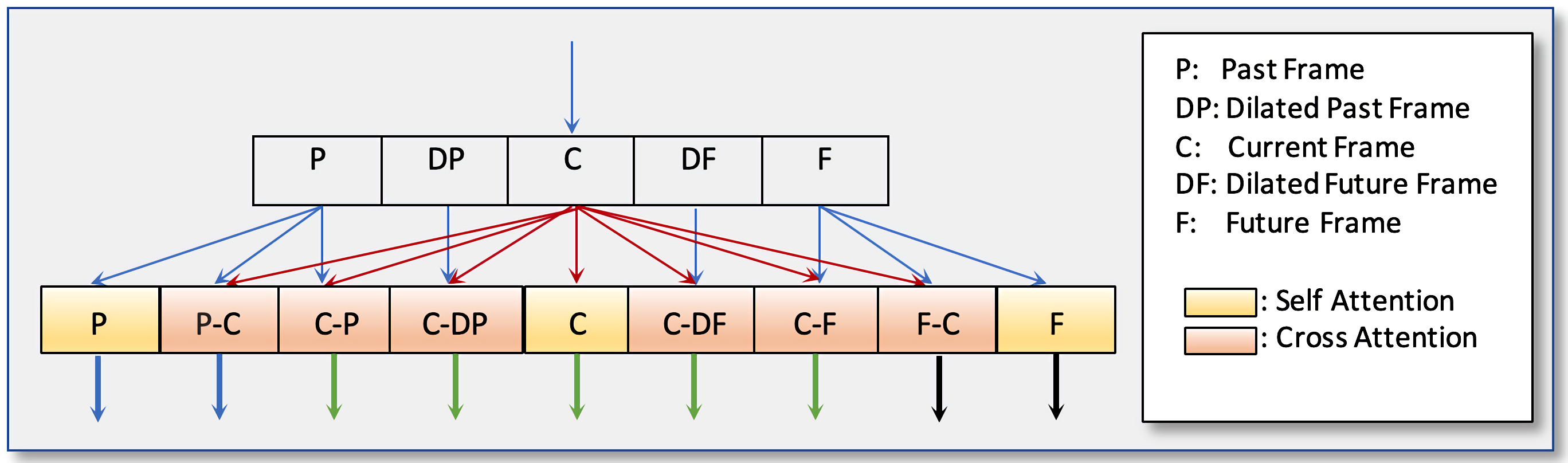}}}
\caption{Multiple Self-Cross Dilated Attention. We
get past,current and future frame feature maps for next stage by fusing attentions of all blue, green and black color arrows separately.}
\label{fig:totalattn}
\end{figure}


We use the cross and dilated cross attention modules to output different attention maps for $\mathbf{f}_i^{t-1}$, $\mathbf{f}_i^{t}$ and $\mathbf{f}_i^{t+1}$. We illustrate all the feature maps in Figure \ref{fig:totalattn}. 
For $\mathbf{f}_{i}^t$, we compute five attention maps in total as $CA^{i}(t,t-1)$, $DCA^{i}(t,t-1)$, $CA_{i}(t,t+1)$, $DCA_{i}(t,t+1)$ and $SA_i(t)$. For $\mathbf{f}_{i}^{t-1}$, we compute $SA_{i}(t-1)$ and $CA_{i}(t-1,t)$, and for  $\mathbf{f}_{i}^{t+1}$, we compute $SA_{i}(t+1)$ and $CA_{i}(t+1,t)$. 
We fuse these attention maps through convex combinations where the weights corresponding to a given attention map are computed by passing them through a convolutional layer and taking softmax. We explain this through a figure in the supplementary. For example, we output weight maps corresponding to each attention map $CA^{i}(t,t-1)$, $DCA^{i}(t,t-1)$, $CA_{i}(t,t+1)$, $DCA_{i}(t,t+1)$ and $SA_i(t)$ through a convolutional layer and apply softmax on the output to obtain weights and combine the attention maps. Thus, the fusion process is adaptive as the weights depend on the corresponding attention. 
The output of fusion module is further processed with RCAB \cite{RCAB} to produce $f^t_{i+1}$. 

After completion of all multi-stages of encoder and the bottleneck, for $N$ stages, we get $\{f^{t-1}_{N+1}, f^{t}_{N+1}, f^{t+1}_{N+1}\}$. Since all the information across frames has been mixed in various encoders, we only use the current frame feature map $f^{t}_{N+1}$ and process it further in the decoder. Thus the decoder architecture remains the same as discussed in the Base Model. 
After the final stage of the decoder, we apply a single convolutional layer to get a temporally stable enhanced frame.

\label{chap:Experiments}
\section{Experiments \& Results}
\subsection{Experimental Setup}\label{Experimental Setup}
We perform experiments by training on static video datasets and testing on datasets with dynamic videos. In particular we adopt three experimental settings. 

\textbf{Setting 1:} We train on static RGB videos obtained from the DRV Dataset \cite{SMID} and evaluate on the dynamic videos. In particular, since the low light videos are available in RAW format, we convert them to RGB using the python library \emph{libraw} and reduce the spatial resolution to 832 $\times$ 1248. There are 202 low light static video sequences each with 110 frames for which the long exposure ground truth is available. We train on 153 videos similar to \cite{SMID}.  We test on 22 dynamic low light videos for which no ground truth is available. This is the most realistic setting that one encounters in the real-world.  

\textbf{Setting 2:} We generate 153 static low light videos using frames from the videos in the DAVIS dataset \cite{DAVIS}. In particular, we apply a gamma transform on a given frame and add multiple instances of noise similar to \cite{Learning_temporal} to generate a static video sequence. We then test on 30 synthetically generated low light dynamic videos from the DAVIS test dataset obtained similar to \cite{Learning_temporal}. We note that while the motion in the above videos is realistic, the distortions are synthetically generated. 

\textbf{Setting 3:} In this setting, we consider realistic distortions but synthetic motion. In particular, we introduce camera motion in the static videos in the DRV dataset described in Setting 1. We estimate the depth of the videos using Midas \cite{MIDAS} and apply different camera trajectories from KITTI camera poses \cite{Kitti} and VEED \cite{Vijaylakshmi} to generate videos with motion. Since some of the generated videos can contain disocclusions, we select a subset of 96 videos without such artifacts for testing. Each video contains 10 frames of resolution 832 $\times$ 1248. We refer to this dataset as the DRV Dataset with Synthetic Motion (DRV-SM). We use the static videos described in Setting 1 for training. 

We note that ground truth videos are available for performance evaluation in Setting 2 and 3. We evaluate the methods using peak signal to noise ratio (PSNR), structural similarity index (SSIM) \cite{SSIM} and spatio-temporal entropic differences (ST-RRED) \cite{ST-RRED}. For Setting 1, we evaluate through a subjective study.  

\subsection{Implementation Details}
While training on static videos, we select three consecutive frames from each sequence. We choose a spatial patch size of 384 $\times$ 384. We train our model with a combination of VGG loss \cite{VGG} and mean squared error loss and use Adam Optimizer \cite{ADAM}. In all the settings, we train our model for 900 epochs; where we use a learning rate of 1e-4. The batch size is set to 2. 
We implement our architecture using PyTorch and use an NVIDIA DGX Version 4.6.0 GPU with 32 GB of memory to train our model. 

\subsection{Performance Evaluation and Comparison}
\begin{figure}[h]

\begin{minipage}[h]{.16\linewidth}
  \centering
  \centerline{\includegraphics[width=\linewidth]{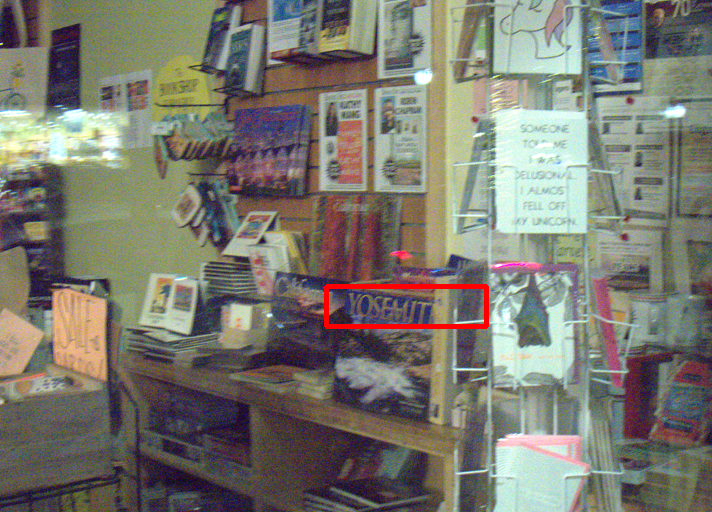}}
\end{minipage}
\hfill
\begin{minipage}[h]{.16\linewidth}
  \centering
  \centerline{\includegraphics[width=\linewidth]{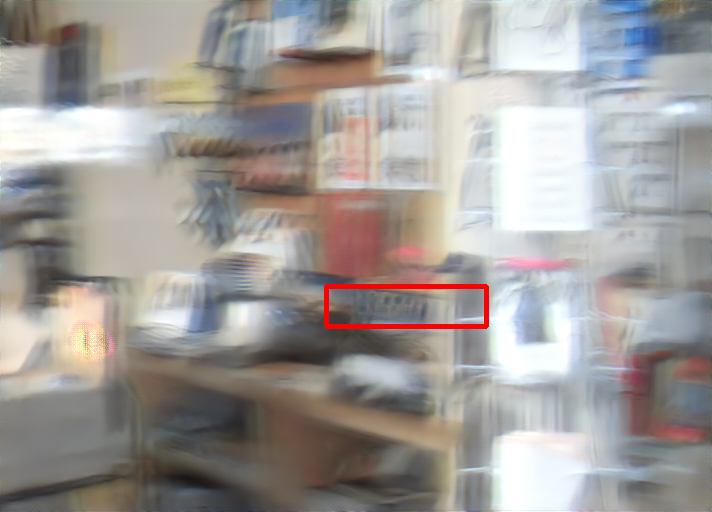}}
\end{minipage}
\hfill
\begin{minipage}[h]{.16\linewidth}
  \centering
  \centerline{\includegraphics[width=\linewidth]{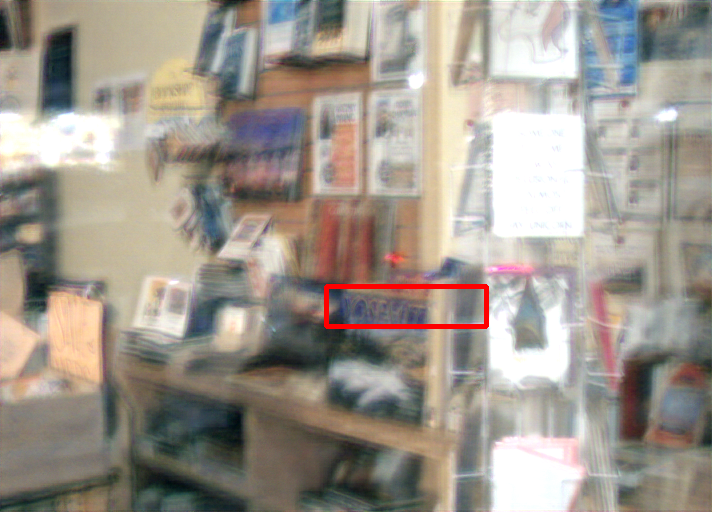}}
\end{minipage}
\hfill
\begin{minipage}[h]{.16\linewidth}
  \centering
  \centerline{\includegraphics[width=\linewidth]{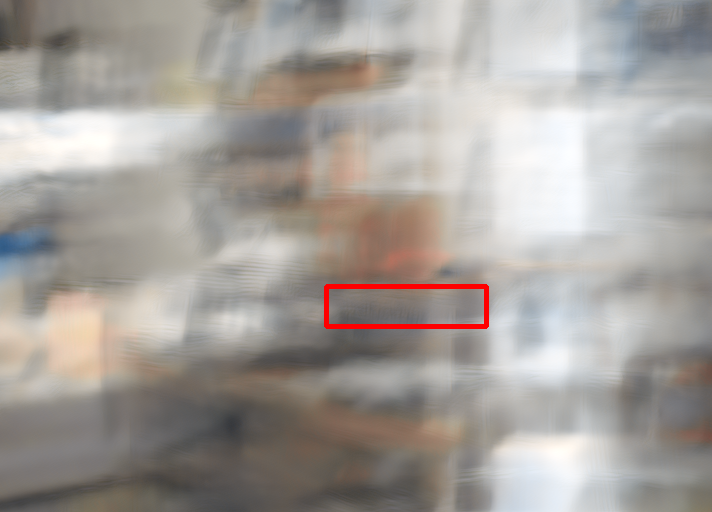}}
\end{minipage}
\hfill
\begin{minipage}[h]{.16\linewidth}
  \centering
  \centerline{\includegraphics[width=\linewidth]{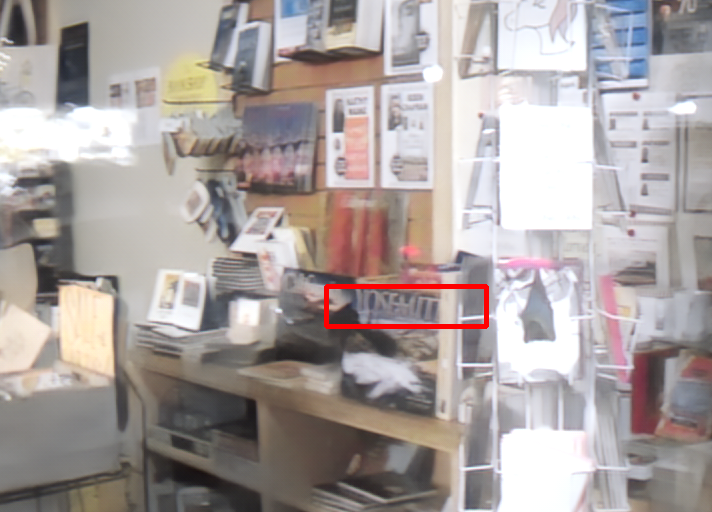}}
\end{minipage}
\hfill
\begin{minipage}[h]{.16\linewidth}
  \centering
  \centerline{\includegraphics[width=\linewidth]{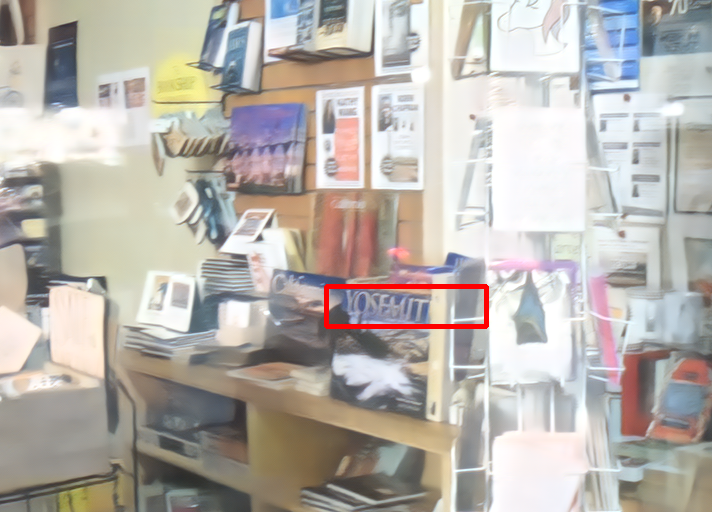}}
\end{minipage}
\\

\begin{minipage}[h]{.16\linewidth}
  \centering
  \centerline{\includegraphics[width=\linewidth]{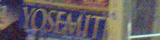}}
   \centerline{Low-Light}\medskip
\end{minipage}
\hfill
\begin{minipage}[h]{.16\linewidth}
  \centering
  \centerline{\includegraphics[width=\linewidth]{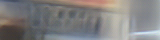}}
\centerline{FastDVDNet}\medskip
\end{minipage}
\hfill
\begin{minipage}[h]{.16\linewidth}
  \centering
  \centerline{\includegraphics[width=\linewidth]{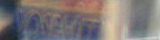}}
  \centerline{VRT}\medskip
\end{minipage}
\hfill
\begin{minipage}[h]{.16\linewidth}
  \centering
  \centerline{\includegraphics[width=\linewidth]{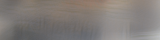}}
  \centerline{MBLLVEN}\medskip
\end{minipage}
\hfill
\begin{minipage}[h]{.16\linewidth}
  \centering
  \centerline{\includegraphics[width=\linewidth]{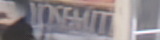}}
  \centerline{SMID}\medskip
\end{minipage}
\hfill
\begin{minipage}[h]{.16\linewidth}
  \centering
  \centerline{\includegraphics[width=\linewidth]{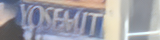}}
  \centerline{Ours}\medskip
\end{minipage}
\\
\begin{minipage}[h]{.137\linewidth}
  \centering
  \centerline{\includegraphics[width=\linewidth]{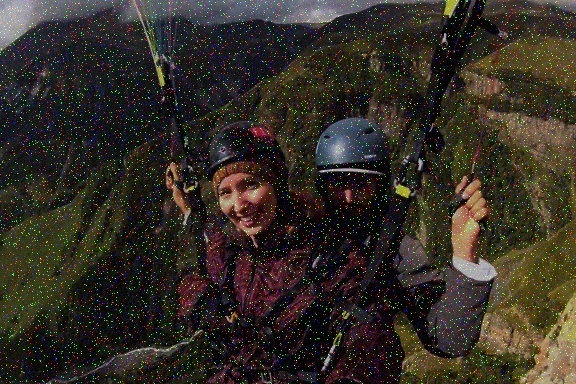}}
\end{minipage}
\hfill
\begin{minipage}[h]{.137\linewidth}
  \centering
  \centerline{\includegraphics[width=\linewidth]{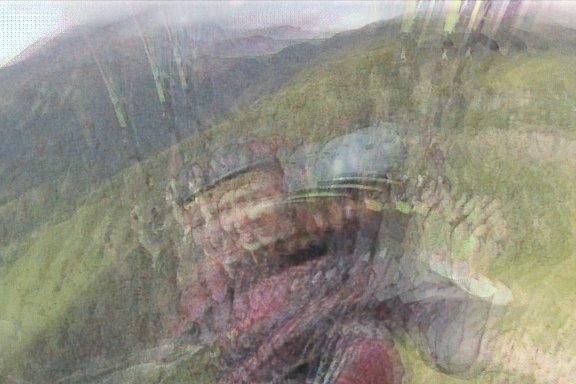}}
\end{minipage}
\hfill
\begin{minipage}[h]{.137\linewidth}
  \centering
  \centerline{\includegraphics[width=\linewidth]{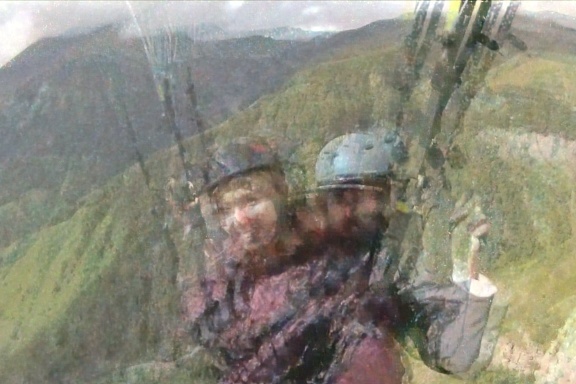}}
\end{minipage}
\hfill
\begin{minipage}[h]{.137\linewidth}
  \centering
  \centerline{\includegraphics[width=\linewidth]{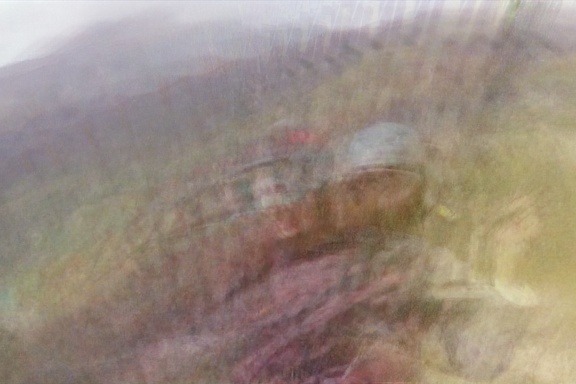}}
\end{minipage}
\hfill
\begin{minipage}[h]{.137\linewidth}
  \centering
  \centerline{\includegraphics[width=\linewidth]{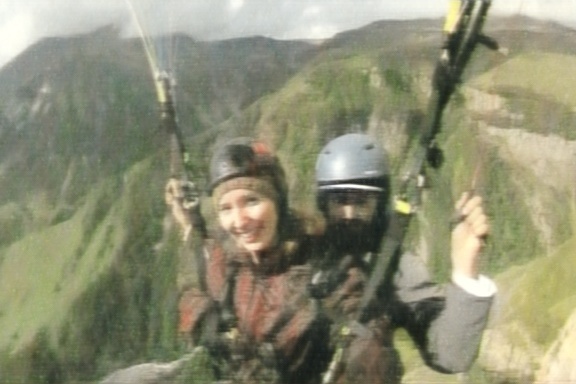}}
\end{minipage}
\hfill
\begin{minipage}[h]{.137\linewidth}
  \centering
  \centerline{\includegraphics[width=\linewidth]{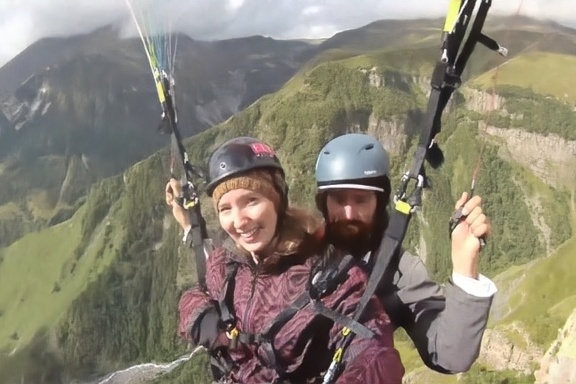}}
\end{minipage}
\hfill
\begin{minipage}[h]{.137\linewidth}
  \centering
  \centerline{\includegraphics[width=\linewidth]{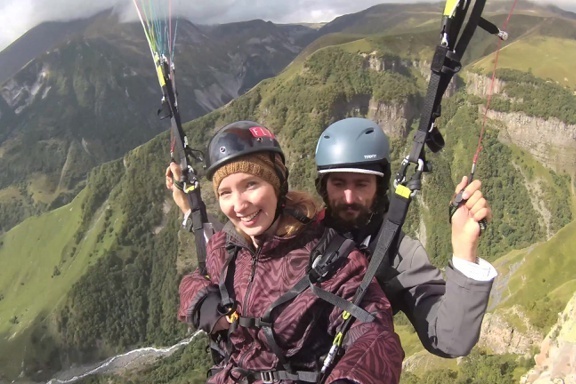}}
\end{minipage}
\\
\begin{minipage}[h]{.137\linewidth}
  \centering
  \centerline{\includegraphics[width=\linewidth]{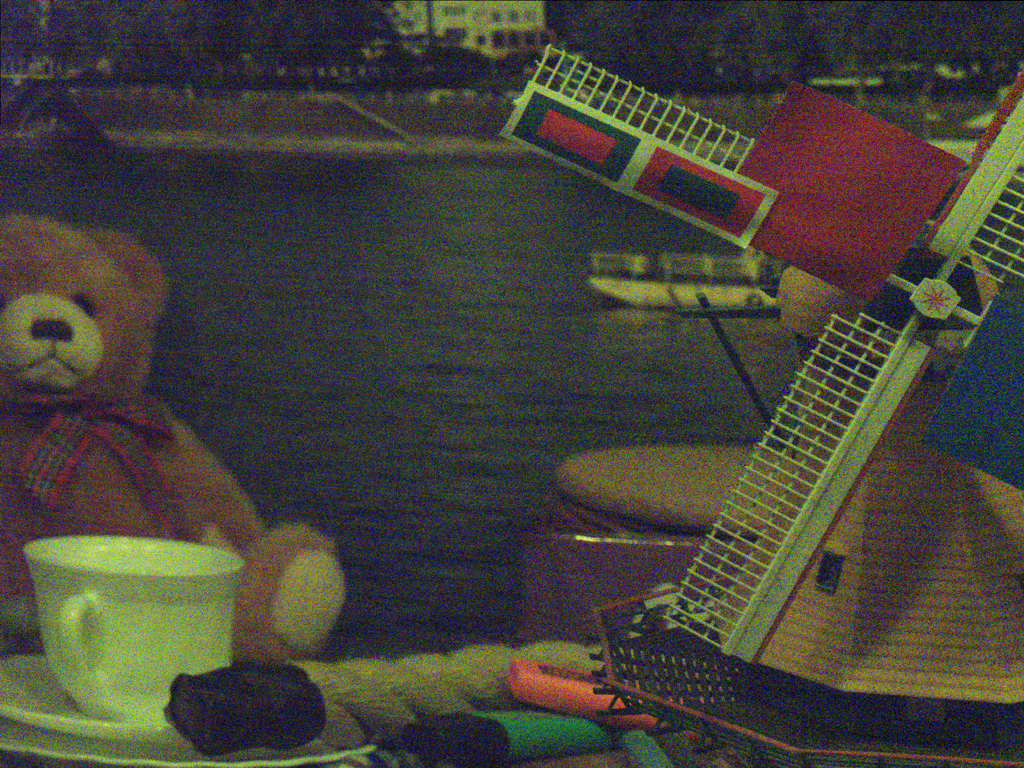}}
  \centerline{Low-Light}\medskip
\end{minipage}
\hfill
\begin{minipage}[h]{.137\linewidth}
  \centering
  \centerline{\includegraphics[width=\linewidth]{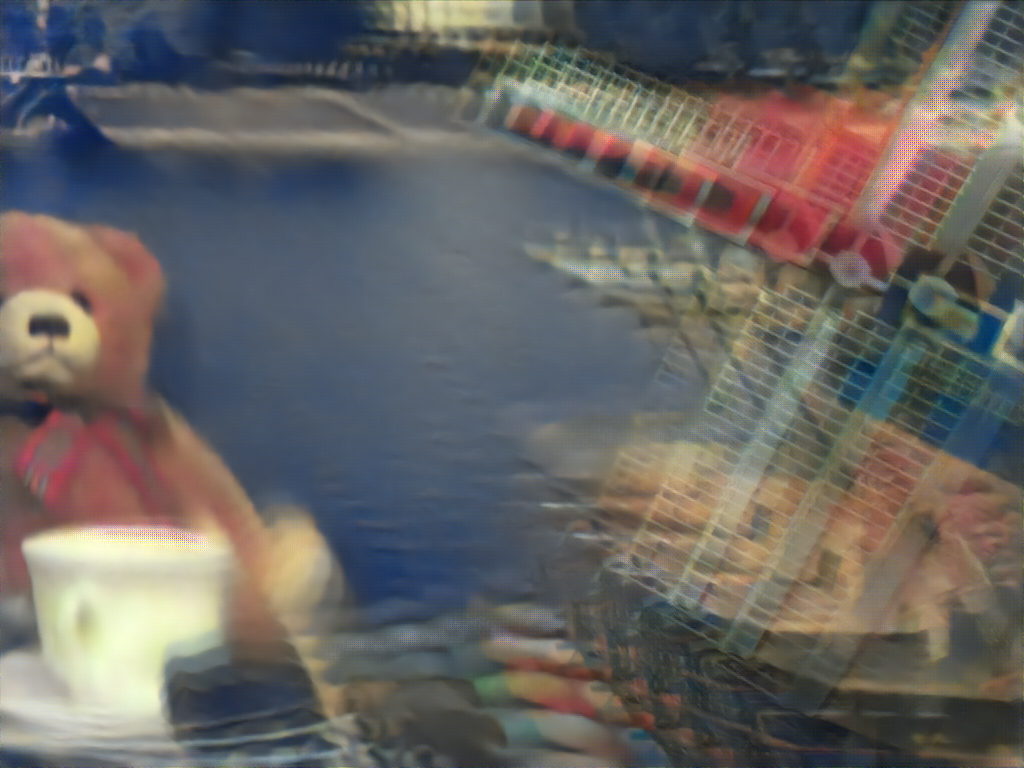}}
\centerline{FastDVDNet}\medskip
\end{minipage}
\hfill
\begin{minipage}[h]{.137\linewidth}
  \centering
  \centerline{\includegraphics[width=\linewidth]{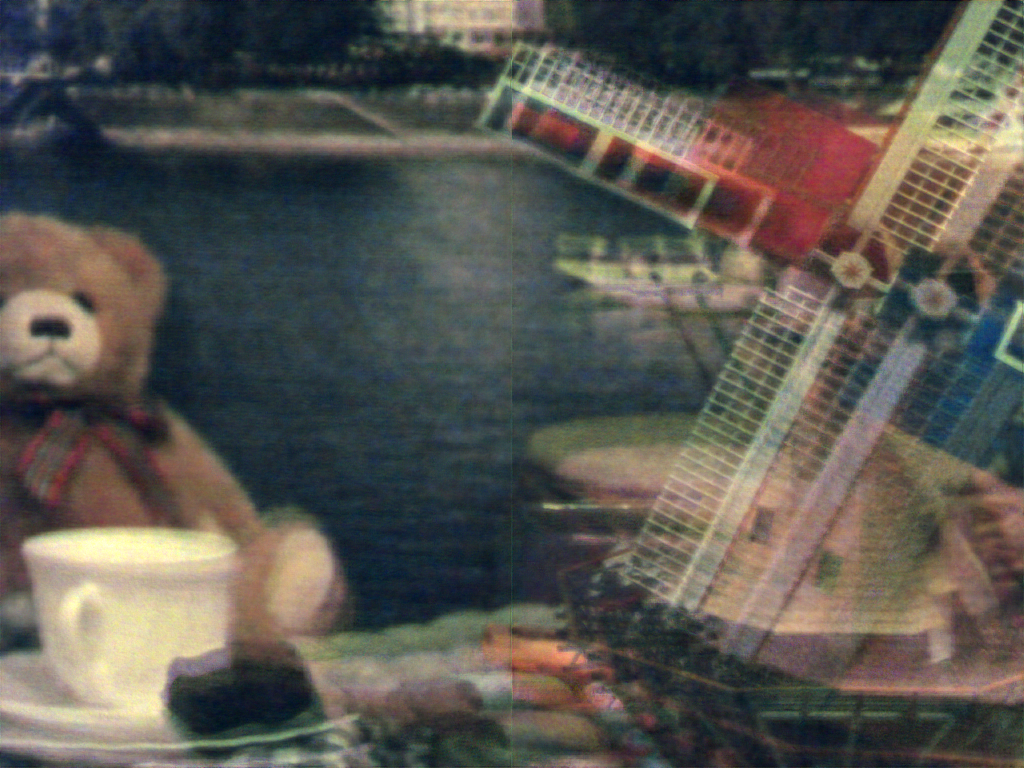}}
  \centerline{VRT}\medskip
\end{minipage}
\hfill
\begin{minipage}[h]{.137\linewidth}
  \centering
  \centerline{\includegraphics[width=\linewidth]{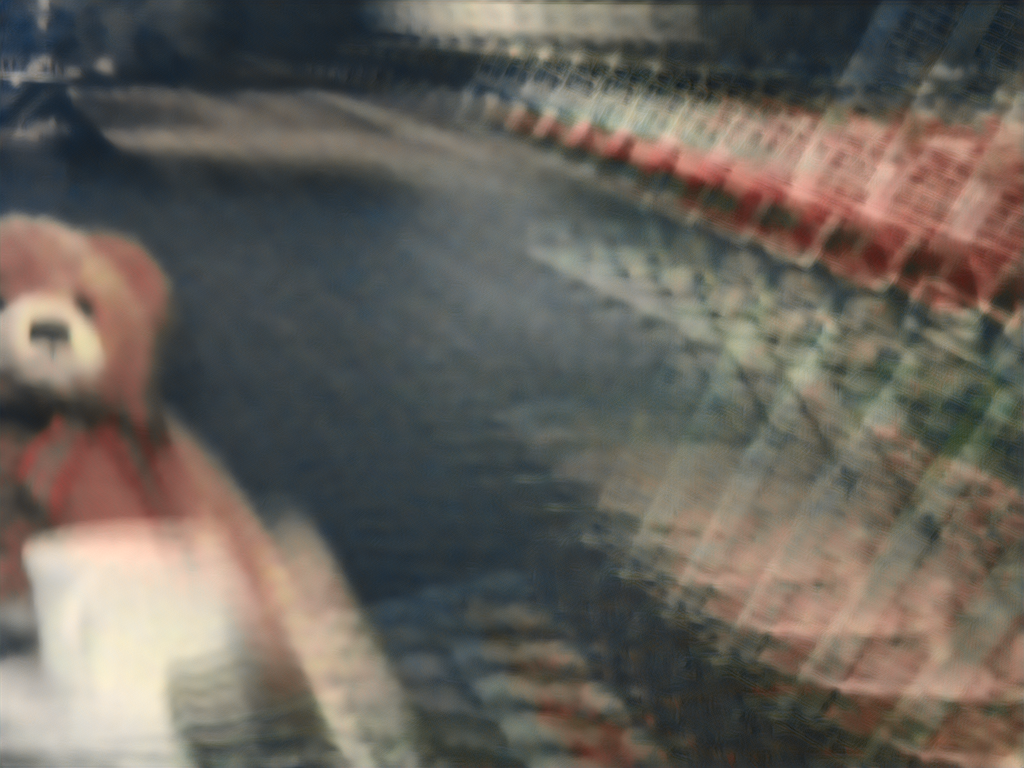}}
  \centerline{MBLLVEN}\medskip
\end{minipage}
\hfill
\begin{minipage}[h]{.137\linewidth}
  \centering
  \centerline{\includegraphics[width=\linewidth]{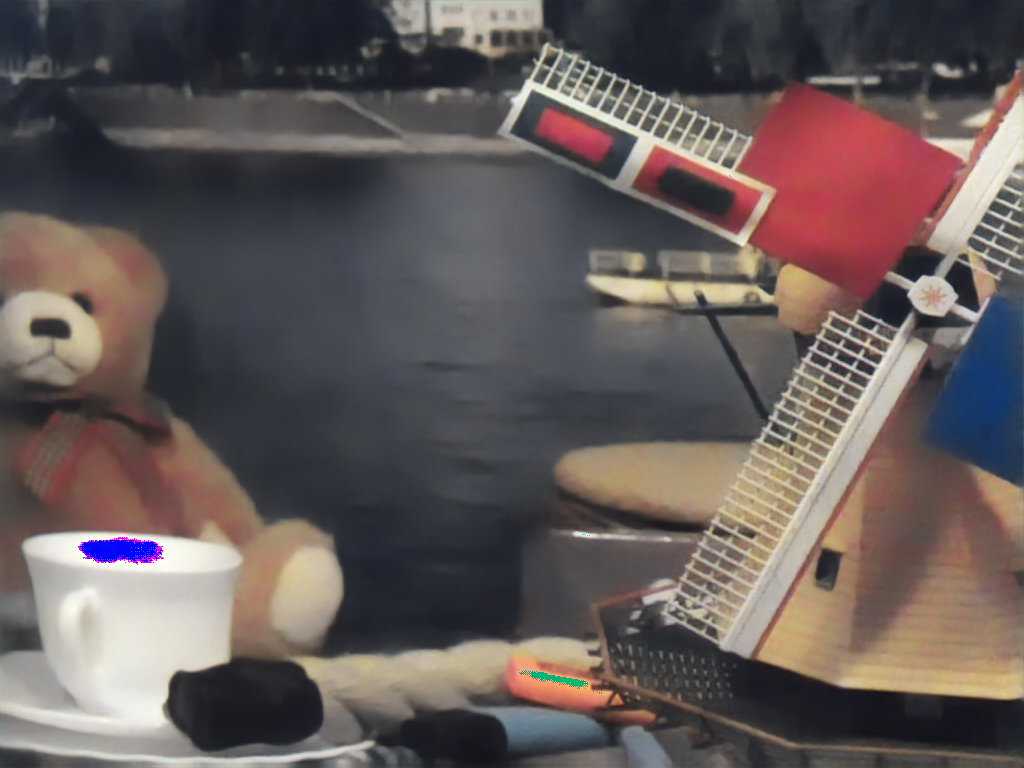}}
  \centerline{SMID}\medskip
\end{minipage}
\hfill
\begin{minipage}[h]{.137\linewidth}
  \centering
  \centerline{\includegraphics[width=\linewidth]{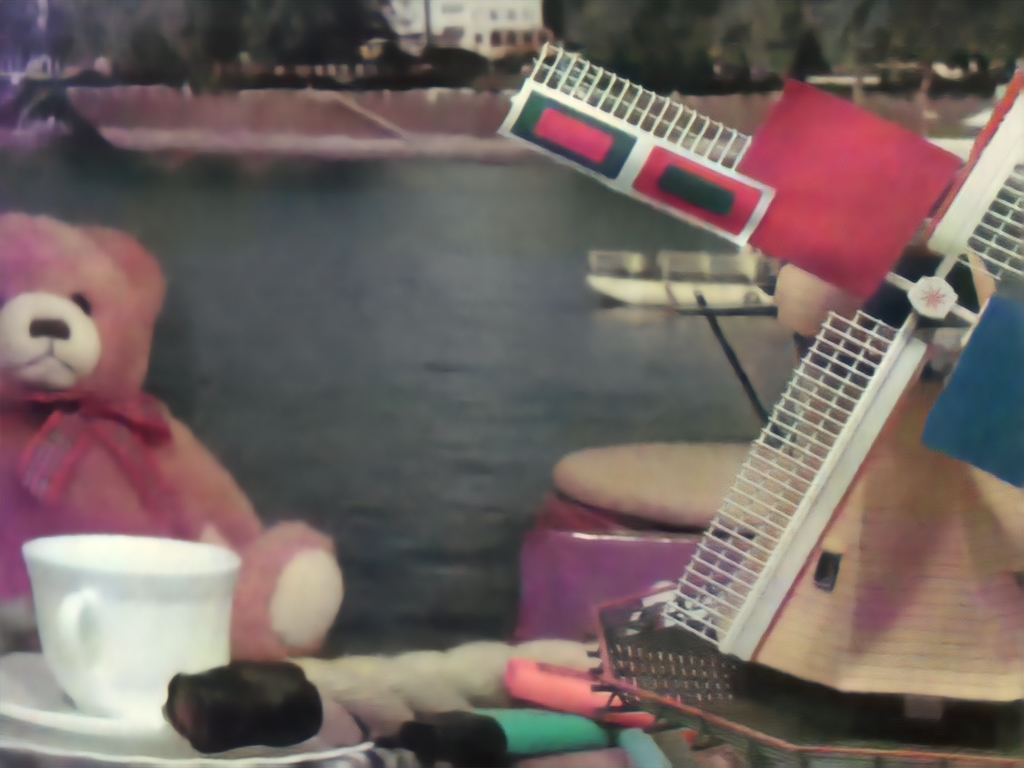}}
  \centerline{Ours}\medskip
\end{minipage}
\hfill
\begin{minipage}[h]{.137\linewidth}
  \centering
  \centerline{\includegraphics[width=\linewidth]{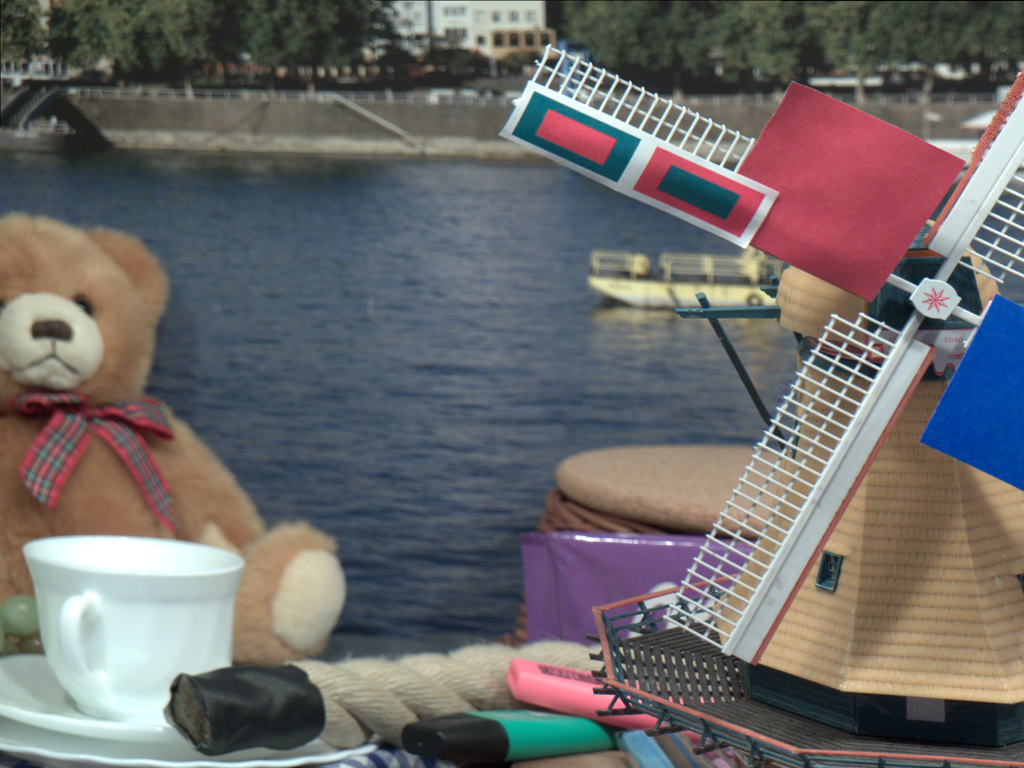}}
  \centerline{Ground Truth}\medskip
\end{minipage}

\caption{Example frames of low-light videos enhanced using various methods. The videos in the $1^{st}$, $2^{nd}$ and $3^{rd}$ rows correspond to Setting 1, 2 and 3 as described in Section \ref{Experimental Setup}. Note that the enhanced frames from our method are sharp and have better perceptual quality when compared to SMID \cite{SMID}. Zoom-in for better viewing.}
\label{Fig:Visual_Performance}
\end{figure}

We compare with state of the art low light video enhancement methods such as SMID \cite{SMID} and MBLLVEN \cite{MBLLEN}. We compare with SMID by adopting the same training method based on consistency of enhanced output for restoring RGB video frames.  We also compare with some of the recent video restoration methods such as FastDVDnet \cite{FastDVDnet} and VRT \cite{VRT}. We discuss issues with comparison of low light video enhancement methods based on data generation in the supplementary. 
All these competing methods are trained and tested similar to our approach as described in Settings 1, 2 and 3. 
We first present visual examples of our results in Figure \ref{Fig:Visual_Performance} corresponding to each of the experimental settings. We clearly see that our method outperforms all the other methods. In particular, generic video restoration methods fail when they are trained on static videos and tested on dynamic videos. While SMID \cite{SMID} performs better, our method outputs even better results with superior enhancement and clear visibility of details. 


\begin{figure}[h]
\begin{tabular}{cc}
\bmvaHangBox{{\includegraphics[width=6cm]{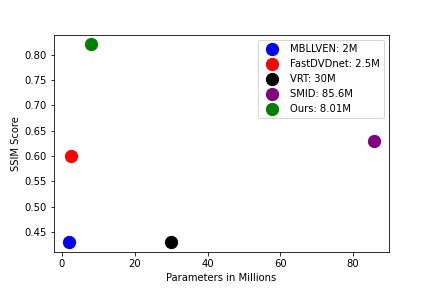}}}&
\bmvaHangBox{{\includegraphics[width=6cm]{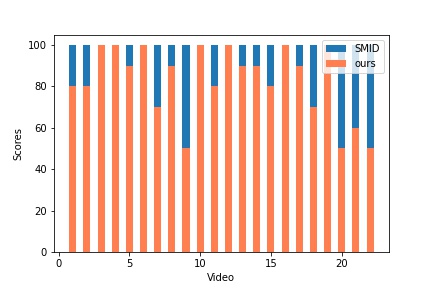}}}\\
(a)&(b)
\end{tabular}
\caption{(a) Number of Parameters vs SSIM, (b) Subjective Study on Dynamic DRV Dataset (Setting 1)}
\label{fig:Parameters_subjective}
\end{figure}

In Table \ref{DAVIS Results}, we present the numerical comparisons on the DAVIS (Setting 2) and DRV-SM datasets (Setting 3), since a ground truth video is available. 
We see that our method performs better than the other methods for both the settings corresponding to realistic motion and realistic low light distortions. We also see in Figure \ref{fig:Parameters_subjective}(a) that our model achieves a very good performance with very few parameters when compared to other methods. 

\begin{table}[h]
\caption{Quantitative Evaluation for Setting 2 and 3} 
\centering 
\begin{tabular}{c|c@{\hskip -0.01in}c@{\hskip -0.01in}c|c@{\hskip -0.01in}c@{\hskip -0.01in}c} 

\hline\hline 
Methods & & Setting 2 (DAVIS) & & & Setting 3 (DRV-SM)\\
\cline{2-7}
& SSIM ${\uparrow}$ & ST-RRED${\downarrow}$ & PSNR${\uparrow}$ & SSIM${\uparrow}$ & ST-RRED${\downarrow}$ & PSNR${\uparrow}$ \\ [1ex] 
\hline 
SMID \cite{SMID}   &0.63 & 682 & 28.63 & 0.55 & 1248 & 28.64\\
MBLLVEN \cite{MBLLEN}   & 0.43 & 2679 & 28.09 & 0.50 & 3628 & 28.38\\
FastDVDNet \cite{FastDVDnet}  & 0.60 & 1624 & 28.36 & 0.54 & 2317 & 28.45\\
VRT \cite{VRT} & 0.43 & 955 & 27.85 & 0.52 & 1882 & 28.47\\
\hline
Ours  & \textbf{0.82} & \textbf{241} & \textbf{29.02} & \textbf{0.60} & \textbf{745} & \textbf{28.92}\\[1ex] 
\hline 
\end{tabular}
\label{DAVIS Results} 
\end{table}

\textbf{Subjective Evaluation on Dynamic DRV Dataset:}
We evaluate the results on the dynamic videos of the DRV dataset through a pairwise subjective study since the ground truth videos are not available. In particular, we compare the performance of our method against SMID, which is the second best approach in Table \ref{DAVIS Results}. The subjective study involved 10 subjects who compared these methods on an LG (27 Inch) IPS Monitor. Our results in Figure \ref{fig:Parameters_subjective}(b) indicate that out of 22 dynamic video sequences, our approach was rated as better than SMID on 19 videos and on the remaining 3, the ratings were equally split between our model and SMID.

\subsection{Ablation Study}
\begin{figure}[h]
\begin{tabular}{cc}
\bmvaHangBox{{\includegraphics[width=6cm]{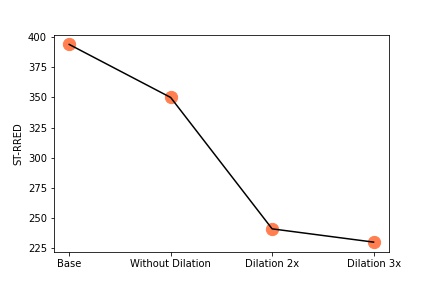}}}&
\bmvaHangBox{{\includegraphics[width=6cm]{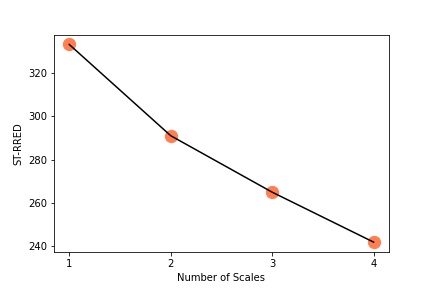}}}\\
(a)&(b)
\end{tabular}
\caption{(a) Ablation evaluation of our models, (b) Performance variation with number of encoders}
\label{fig:Ablation}
\end{figure}

We evaluate the importance of various components of our model in Figure \ref{fig:Ablation}(a) corresponding to Setting 2 using ST-RRED. We see that our final model improves on both the baseline model and the model without dilation. This shows that the cross-attention module helps account for the motion although it is only trained on static sequences. Further, the dilated attention maps help account for larger motion between neighboring frames. We also see a minor improvement as we increase the dilation factor from 2 to 3. We also evaluate the performance variation with respect to the number of stages corresponding to the number of encoders in our model. We see in Figure \ref{fig:Ablation}(a) that the performance improves with the number of encoders and the gains tend to decrease with more scales. Although the performance could improve with more scales, we limit to 4 scales due to memory constraints.  

\subsection{Performance on Static DRV Dataset}
Although our primary objective was in the performance evaluation of the dynamic videos, here we also evaluate the performance of our method on the static videos in the DRV dataset referred to as Static DRV dataset \cite{SMID}. The models are all trained on the same static sequences mentioned in Setting 1, but now evaluated on the test set of Static DRV dataset consisting of 49 video sequences. In particular, we test on the first 10 frames from each of these static sequences in Table \ref{DDRV Results}. We evaluate using SSIM and PSNR since there is no motion in the video sequences. We note that there is no clear winner with respect to different performance measures and our approach is competitive with the best. To summarize, our model matches the performance of other methods on static sequences, yet achieves a significantly better performance on dynamic videos. 

\begin{table}[h]
\label{DDRV Results}
\caption{Performance On Static DRV Dataset} 
\centering 
\begin{tabular}{c c c} 
\hline\hline 

Methods  & SSIM ${\uparrow}$ & PSNR ${\uparrow}$\\ [1ex]
\hline 
SMID\cite{SMID}   & \textbf{0.63}  & 28.43\\
MBLLVEN\cite{MBLLEN}   & 0.58  & 28.36\\
FastDVDNet\cite{FastDVDnet}  & 0.62 & 28.52\\
VRT\cite{VRT}  & 0.53 & 28.37\\
Ours  & 0.61 & \textbf{28.53}\\[1ex] 
\hline 
\end{tabular}
\end{table}

\section{Conclusion}
We explored the use of attention based modules for learning of low light video enhancement on static videos for their application on dynamic videos. We showed that these attention modules can obviate the need for explicit optical flow estimation yet account for inter-frame interactions even though the interaction dynamics are different between training and testing. We achieve superior performance than other methods on multiple datasets. Our approach may also be relevant for generic video restoration.

\newpage
\pagebreak
\begin{center}
\textbf{\large Supplemental Materials}
\end{center}
\label{chap:Supplementary}
\section{Architecture of Fusion Module}

\label{sec:Fusion}
\begin{figure}[h]
\centering
\bmvaHangBox{{\includegraphics[width=12.4cm]{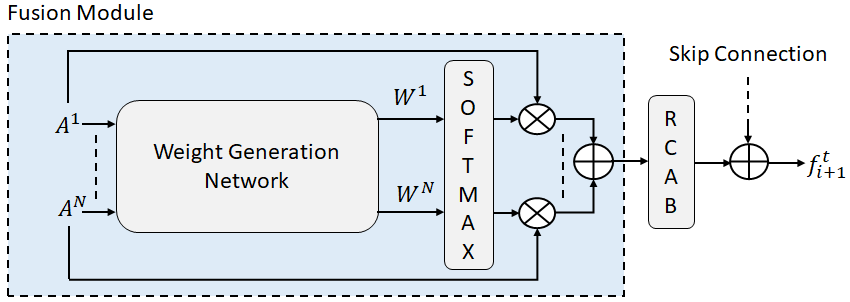}}}
\caption{Block diagram illustrating our fusion module}
\label{fig:fusion}
\end{figure}
In this section, we describe the adaptive fusion for $N$ attention maps (see Figure \ref{fig:fusion}), $\{A^1, A^2, \ldots, A^N\}$.
We pass each of these maps through a convolutional layer and get $W^n$ for $n=1,2,\ldots,N$, where each $W^n$ is a single channel map. The same map will be used to determine the weights for all the channels in the attention map. Finally, to obtain the mixing weights for $A^n$ at each location, we apply a softmax operation at each spatial location across all the $N$ maps. The resulting weights are used to fuse the attention map followed by the residual channel attention block (RCAB). 





\section{Ablation on DRV-SM Dataset}\label{sec:Ablation}
We evaluated the importance of various components of our model, in particular, cross-attention and dilated cross-attention with respect to our base line model on the DAVIS dataset in the main paper. Here we present similar results on the DRV-SM dataset using the ST-RRED quality measure in Figure \ref{BaseAttnModule}. We see consistent improvements with respect to our ideas even in the DRV-SM dataset.  

\begin{figure}[h]
\label{BaseAttnModule}
\centering
\bmvaHangBox{{\includegraphics[width=6.2cm]{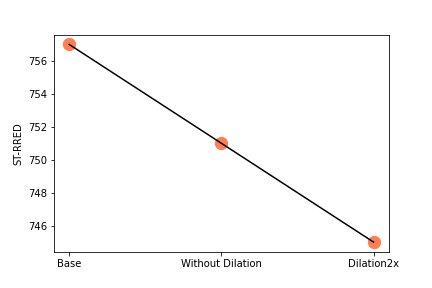}}}
\caption{Ablation Evaluation on our model with DRV-SM dataset}
\end{figure}

\section{Comparison with Data Generation based Methods}
\begin{figure*}[h]
\begin{center}
  \begin{minipage}[b]{0.245\linewidth}
  \centering
  \centerline{\includegraphics[width=\linewidth]{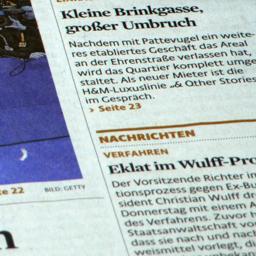}}
\end{minipage}
\hfill
\begin{minipage}[b]{0.245\linewidth}
  \centering
  \centerline{\includegraphics[width=\linewidth]{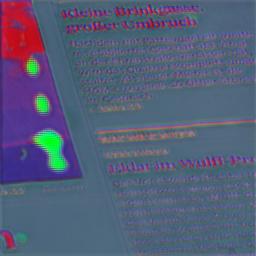}}
\end{minipage}
\hfill
\begin{minipage}[b]{0.245\linewidth}
  \centering
  \centerline{\includegraphics[width=\linewidth]{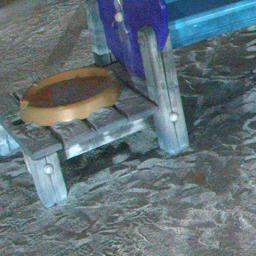}}
\end{minipage}
\hfill
\begin{minipage}[b]{0.245\linewidth}
  \centering
  \centerline{\includegraphics[width=\linewidth]{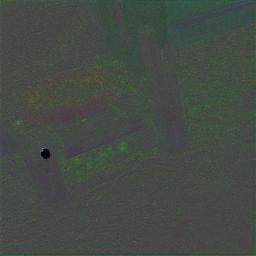}}
\end{minipage}
\\
  \begin{minipage}[b]{0.245\linewidth}
  \centering
  \centerline{\includegraphics[width=\linewidth]{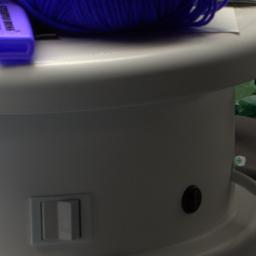}}
\end{minipage}
\hfill
\begin{minipage}[b]{0.245\linewidth}
  \centering
  \centerline{\includegraphics[width=\linewidth]{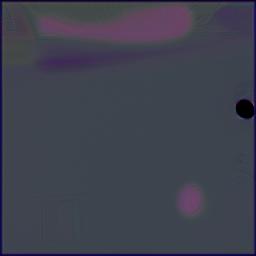}}
\end{minipage}
\hfill
\begin{minipage}[b]{0.245\linewidth}
  \centering
  \centerline{\includegraphics[width=\linewidth]{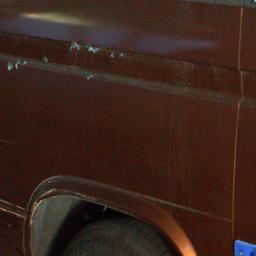}}
\end{minipage}
\hfill
\begin{minipage}[b]{0.245\linewidth}
  \centering
  \centerline{\includegraphics[width=\linewidth]{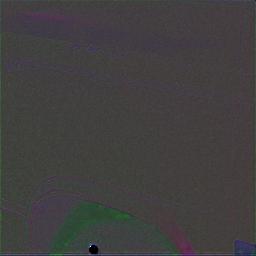}}
\end{minipage}

\caption{Examples of images generated using the SIDGAN \cite{CycleGAN}. First column are the ground truth images from the Vimeo dataset. Second column contains outputs from first stage of SIDGAN of transforming Vimeo ground truth to sensor specific ground truths from the DRV dataset. Third column contains ground truth images from the DRV dataset and the last column contains outputs of the second stage of the SIDGAN of transforming images from third column to low light from the DRV dataset. Note that the generated images appear unnatural.}
\label{fig:analysis}
\end{center}
\label{fig:sidgan}
\end{figure*}

We now describe our experiments in comparing with a data generation method for training low light video enhancement \cite{CycleGAN}. SIDGAN adopts a two-step procedure to generate realistic low light videos from high quality videos in the Vimeo dataset \cite{vimeo}. In particular, a pair of cycle-GANs is used where the first cycle-GAN generates sensor specific long exposure video frames from which a second cycle-GAN generates short exposure video frames. Thus, paired labelled data is created which can be used to train deep network. However, we observe that each cycle-GAN generates very poor quality images in our experiments as shown in Figure \ref{fig:sidgan}. Thus we did not proceed with training the enhancement model. We believe that poor quality of video frames could perhaps be attributed to the larger image resolutions that we operate with when compared to SIDGAN \cite{CycleGAN}. Due to memory constraints, we are not able to experiment with larger patch sizes in their model which could potentially improve performance. 

\bibliography{egbib}
\end{document}